\documentclass{article}

\usepackage[preprint]{neurips_2026}

\usepackage[utf8]{inputenc} % allow utf-8 input
\usepackage[T1]{fontenc}    % use 8-bit T1 fonts
\usepackage{hyperref}       % hyperlinks
\usepackage{url}            % simple URL typesetting
\usepackage{booktabs}       % professional-quality tables
\usepackage{amsfonts}       % blackboard math symbols
\usepackage{nicefrac}       % compact symbols for 1/2, etc.
\usepackage{microtype}      % microtypography
\usepackage{xcolor}         % colors
\usepackage{amsmath}
\usepackage{graphicx}
\usepackage{makecell}
\usepackage{enumitem}

\title{Diffusion-APO: Trajectory-Aware Direct Preference Alignment for Video Diffusion Transformers}

\author{%
  Jingyuan Zhu$^{1}$, Biaolong Chen$^{1}$, Le Zhang$^{1}$, Aixi Zhang$^{1}$, Hao Jiang$^{1}$, Pipei Huang$^{1}$ \\
  $^{1}$Alibaba Group \\
}

\begin{document}

\maketitle

\begin{abstract}
Efficiently aligning large-scale video diffusion models with human intent requires a scalable and trajectory-aware pathway that bridges the inherent discrepancy between training noise distributions and practical inference trajectories. While existing paradigms such as Direct Preference Optimization (DPO) and Group Relative Policy Optimization (GRPO) attempt to address this, they are often hindered by either reliance on bias-prone, complex reward models or suboptimal timestep sampling. In this paper, we propose \textbf{Diffusion-APO} (Aligned Preference Optimization), a trajectory-aware algorithm that resolves this misalignment by synchronizing training noise with inference-time denoising paths to maximize gradient signal efficacy. To translate this algorithmic innovation into a practical solution, we introduce a unified and modular RLHF framework that integrates online ranking, half-online anchoring, offline refinement, and distillation-aware drift correction. This framework enables flexible, multi-stage preference alignment across diverse data and computational constraints without relying on scalar-reward-based policy gradients. Through extensive experiments, we demonstrate that Diffusion-APO consistently outperforms standard baselines in visual quality and instruction following, while effectively preserving generative fidelity during model acceleration, providing a robust, end-to-end pathway for scalable video diffusion alignment.

\end{abstract}

\section{Introduction}
Recent advances in Video Diffusion Models (VDMs)~\cite{ho2022video,blattmann2023align}, particularly the emergence of Video Diffusion Transformers (DiTs)~\cite{peebles2023scalable}, have significantly advanced high-fidelity video synthesis. Scaling model parameters and training data has enabled remarkable capabilities in commercial systems such as Kling~\cite{kling2024kling}, Sora~\cite{sora}, Seedance~\cite{seedance2025seedance}, and Wan~\cite{wan2025wan}. However, aligning pre-trained VDMs with human intent requires more than scale alone. Post-training via Reinforcement Learning from Human Feedback (RLHF)~\cite{bai2022constitutional,ouyang2022training} is essential for mitigating persistent artifacts such as temporal flickering and physical inconsistencies—challenges that supervised fine-tuning on real data inadequately addresses. 

Direct Preference Optimization (DPO) and Group Relative Policy Optimization (GRPO) have emerged as pivotal paradigms for post-training video generation models. However, modern GRPO frameworks critically depend on fine-grained, stable reward models—a dependency that remains a significant bottleneck for video generation, as the lack of mature, open-source video reward models complicates training and introduces severe susceptibility to reward-induced biases. Conversely, while standard DPO operates without an explicit reward model, it suffers from two fundamental limitations: optimization inefficiencies caused by off-policy learning, and a timestep sampling misalignment where noise scheduling is decoupled from the actual inference trajectory.

To address these, this paper proposes Diffusion-APO (Aligned Preference Optimization), a trajectory-aware algorithm that achieves direct preference alignment by resolving the inherent timestep sampling misalignment. By synchronizing training noise distributions with practical inference trajectories, Diffusion-APO maximizes gradient signal efficacy and significantly improves alignment precision compared to standard DPO. Furthermore, our approach bypasses the instability and high computational costs of scalar-reward-based policy gradients, establishing a more stable and efficient alignment path for video diffusion models.

Beyond the core algorithm, we introduce a unified, modular framework that synergizes diverse training paradigms to accommodate varying computational constraints and data availability. Our framework orchestrates a progressive alignment curriculum through: (1) Online DPO, leveraging inference-in-the-loop feedback via a pre-trained ranking model; (2) Half-Online DPO, which balances on-policy exploration with the stability of offline buffers; (3) Offline DPO, providing rigorous final-stage refinement; and (4) Distillation-aware DPO, which mitigates alignment drift under inference acceleration. Crucially, each stage is governed by the consistent Diffusion-APO objective, ensuring theoretical coherence and seamless deployment throughout the entire model lifecycle.

Our main contributions are summarized as follows:

\begin{itemize}
\item \textbf{Trajectory-Aware Alignment:} We introduce \textbf{Diffusion-APO} to resolve the inherent timestep misalignment in existing diffusion-based DPO alignment. By synchronizing training noise distributions with inference trajectories, our method bridges the train-inference gap and maximizes gradient signal efficacy.

\item \textbf{Unified Direct Preference Alignment Framework:} We propose a modular framework that achieves robust alignment while bypassing the instabilities and costs of scalar-reward-based policy gradients. By integrating online, offline, and distillation-aware paradigms, our pipeline provides a scalable, end-to-end solution for aligning video diffusion models under varying computational constraints.

\item \textbf{Empirical Efficacy and Versatility:} Extensive experiments demonstrate that Diffusion-APO consistently outperforms standard DPO in generative fidelity and instruction following. Furthermore, our framework effectively optimizes both baseline models and accelerated versions, ensuring high-quality synthesis across diverse deployment scenarios.
\end{itemize}

\section{Related Work}
\subsection{Diffusion-based Video Generation}
Diffusion models~\cite{ho2020denoisingdiffusionprobabilisticmodels,rombach2022highresolutionimagesynthesislatent,song2021scorebasedgenerativemodelingstochastic} have recently emerged as a dominant paradigm for generative modeling and have demonstrated remarkable success in image and video synthesis. Early works~\cite{blattmann2023align,guo2024animatediffanimatepersonalizedtexttoimage,khachatryan2023text2videozerotexttoimagediffusionmodels,singer2022makeavideotexttovideogenerationtextvideo,wang2023modelscopetexttovideotechnicalreport,wang2023laviehighqualityvideogeneration,wu2023tuneavideooneshottuningimage,zhou2023magicvideoefficientvideogeneration} on video diffusion models introduce temporally-aware architectures that extend image diffusion models to handle video sequences, enabling the generation of coherent short video clips. Subsequent research has significantly improved the scalability and quality of video diffusion models. Imagen video~\cite{ho2022imagenvideohighdefinition} adopts a cascaded diffusion framework to generate high-resolution videos from text prompts. Phenaki~\cite{villegas2022phenakivariablelengthvideo} compresses the video to a small representation of discrete tokens and generates arbitrary long videos conditioned on a sequence of prompts. Subsequent large-scale video diffusion models~\cite{bartal2024lumierespacetimediffusionmodel,blattmann2023stablevideodiffusionscaling,chen2023videocrafter1opendiffusionmodels,chen2024videocrafter2overcomingdatalimitations} show impressive video quality with enriched video datasets. 

More recently, transformer-based diffusion architectures have been widely applied to large-scale video generation. Diffusion Transformers (DiT)~\cite{peebles2023scalable} replace convolutional backbones with transformer blocks, enabling improved scalability and representation capacity for large generative models. Inspired by the breakthrough work SORA~\cite{brooks2024videogenerationworldsimulators}, a series of DiT-based large-scale video diffusion models including ViDu~\cite{bao2024viduhighlyconsistentdynamic}, CogVideo~\cite{hongcogvideo,yangcogvideox}, Wan~\cite{wan2025wan}, Kling~\cite{kling2024kling}, and Seedance~\cite{seedance2025seedance} have demonstrated remarkable capabilities in generating long, high-resolution, and temporally coherent videos given various references. Despite these rapid technical advances, effectively aligning such massive generative models with human intent remains an open challenge. The high dimensionality of video data and the scarcity of scalable, high-quality human feedback complicate the integration of human-centric priors, which often leads to instruction and physical inconsistencies.

\subsection{Preference Alignment for Diffusion Models}
Reinforcement Learning from Human Feedback (RLHF)~\cite{bai2022constitutional,ouyang2022training} has been applied to diffusion models through offline preference learning and online policy optimization. Diffusion-DPO~\cite{wallace2024diffusion} first adapted Direct Preference Optimization (DPO)~\cite{rafailov2023direct} to diffusion models using pairwise human annotations. Diffusion-KTO~\cite{li2024aligning} uses per-image/video binary feedback signals instead of pairwise preference data. OnlineVPO~\cite{zhang2025onlinevpoalignvideodiffusion} extended this with online sampling and periodic reference model updates.

GRPO-based methods have shown particular promise for video diffusion: Flow-GRPO~\cite{liuflow} and Dance-GRPO~\cite{xue2025dancegrpo} apply Group Relative Policy Optimization~\cite{shao2024deepseekmathpushinglimitsmathematical} via SDE sampling in flow matching; MixGRPO~\cite{li2026mixgrpounlockingflowbasedgrpo} streamlines optimization for improved efficiency; and Coefficient-preserving GRPO~\cite{wang2025coefficientspreservingsamplingreinforcementlearning} optimizes the sampling trajectory for visual quality. DiffusionNFT~\cite{zheng2026diffusionnftonlinediffusionreinforcement} introduces an online RL paradigm that optimizes the forward process through contrastive generation. Alternatively, REFL~\cite{xu2023imagerewardlearningevaluatinghuman} fine-tunes models by propagating gradients through reward models, VAEs, and diffusion models---an approach adopted by T2V-Turbo~\cite{li2024t2vturbobreakingqualitybottleneck}, InstructVideo~\cite{yuan2023instructvideoinstructingvideodiffusion}, and VDARG~\cite{prabhudesai2024videodiffusionalignmentreward}.

However, these RM-dependent online methods~\cite{liuflow,xue2025dancegrpo,xu2023imagerewardlearningevaluatinghuman} face a fundamental bottleneck: the lack of robust, consensus-based reward models for video, which often introduces aleatoric uncertainty and exacerbates policy drift. Departing from these approaches, we propose Diffusion-APO, a framework for Direct Preference Alignment. By integrating online and offline DPO directly into the diffusion loop, our method maximizes preference likelihood rather than scalar-reward-based policy gradients. This formulation ensures theoretical consistency and superior deployment efficiency, while effectively bridging the training-inference gap—a limitation overlooked by existing DPO-based solutions.

\section{Method}
In this section, we present our methodology for preference alignment in video diffusion models. We first review preliminaries on diffusion-based video generation, preference optimization, and model distillation in Sec.~\ref{sec31}. We then introduce Diffusion-APO, a trajectory-aware preference optimization algorithm that synchronizes training-time noise distributions with inference trajectories to mitigate optimization mismatch in Sec.~\ref{sec32}. Finally, in Sec.~\ref{sec33}, we present a unified preference optimization framework that leverages Diffusion-APO as its core objective. This modular pipeline integrates online, half-online, offline, and distillation-aware paradigms, enabling scalable and flexible deployment across varying computational constraints and data availability, while bypassing the instabilities of scalar-reward-based policy gradients.

\subsection{Preliminaries}
\label{sec31}
\textbf{Diffusion Models} learn to reverse a gradual noise corruption process. Given a data sample $x_0$, the forward process yields noisy samples $x_t=\sqrt{\bar{\alpha}_t}x_0+\sqrt{1-\bar{\alpha}_t}\epsilon, \quad \epsilon\sim\mathcal{N}(0,\mathbf{I})$ at timestep $t$. The model $\epsilon_\theta(x_t,t,c)$ is trained to predict the added noise conditioned on context $c$ via the standard objective: 
$\mathcal{L}_{\text{DM}} = \mathbb{E}_{x_0,t,\epsilon}\left[\left\|\epsilon-\epsilon_\theta(x_t,t,c)\right\|^2_2\right]$. During inference, the model starts from Gaussian noise $x_T$ and iteratively applies the learned reverse process to generate a clean sample $x_0$.

\textbf{Diffusion-DPO.} Despite the strong generation capabilities of diffusion models, their outputs may not always align with human preferences. Preference learning aims to fine-tune the models using human feedback, typically provided in the form of pairwise comparisons. Given a prompt $c$ and two generated samples $(x^w,x^l)$ where $x^w$ is preferred over $x^l$, DPO optimizes the model by increasing the relative likelihood of the preferred sample without explicitly training a reward model. Specifically, given a reference model $\pi_{ref}$ and a trainable model $\pi_{\theta}$, the DPO objective is defined as:
\begin{equation}
\mathcal{L}_{\text{DPO}} = -\mathbb{E}_{x^w,x^l,c}\left[\log \sigma \left(\beta\left(\log \frac{\pi_\theta(x^w|c)}{\pi_{\text{ref}}(x^w|c)}-\log \frac{\pi_\theta(x^l|c)}{\pi_{\text{ref}}(x^l|c)}\right)\right)\right],
\end{equation}
where $\sigma$ denotes the sigmoid function and $\beta$ controls the strength of preference optimization.

Applying DPO to diffusion models presents a challenge, as evaluating the exact probability density $\pi_{\theta}(x|c)$ is computationally intractable. To bridge this gap, Diffusion-DPO~\cite{wallace2024diffusion} reformulates the required log-likelihood ratios as expectations over sampled timesteps, utilizing the core denoising objective. Given a timestep $t$, the implicit log-probability can be approximated through the noise prediction error: 
\begin{equation}
\log \pi_\theta(x|c) \propto - \mathbb{E}_{t,\epsilon}\left[\left\|\epsilon-\epsilon_\theta(x_t,t,c)\right\|^2_2\right].
\end{equation}
Under this formulation, preference alignment is achieved by contrasting the denoising residuals of chosen and rejected samples across both the active policy and a reference model, seamlessly preserving the standard diffusion framework. However, existing methods typically sample training timesteps independently of the inference trajectory. This introduces a training-inference discrepancy that degrades alignment efficacy—a critical limitation that directly motivates our approach.

\textbf{Diffusion Distillation} seeks to distill the robust generative capabilities of a teacher model into an efficient student architecture. While classifier-free guidance (CFG)~\cite{ho2022classifierfree} significantly enhances condition alignment via the formulation $\epsilon_{\theta}^{\text{cfg}}(x_t,t,c,\omega)=(1+\omega)\epsilon_{\theta}(x_t,t,c)-\omega\epsilon_{\theta}(x_t,t,\emptyset)$, it intrinsically doubles the computational overhead at each generative step. This inefficiency is further exacerbated in video generation, where additional signals like spatiotemporal guidance (STG)~\cite{hyung2024spatiotemporalskipguidanceenhanced} are strictly required to preserve inter-frame motion consistency. To construct our teacher target, we integrate both CFG and STG into a unified guided prediction:
\begin{equation}
\label{eq:cfgstg}
\epsilon_{\theta}^{\text{guided}}(x_t,t,c)=\epsilon_{\theta}^{\text{cfg}}(x_t,t,c,\omega)+\lambda \epsilon_{\theta}^{\text{stg}}(x_t,t,c),
\end{equation}
where $\omega$ and $\lambda$ govern the strengths of the classifier-free and spatiotemporal guidance terms, respectively.

To improve inference efficiency, we distill the composite guidance signal (Eq.~\ref{eq:cfgstg}) into a parameterized student model $\epsilon_{\eta}(x_t,t,c)$, effectively integrating CFG and STG priors into a single forward pass. The objective is defined as:
\begin{equation}
\mathcal{L}_{\text{distill}} =\mathbb{E}_{x_0,t,\epsilon,\omega,\lambda}\left[\left\|\epsilon_{\eta}(x_t,t,c)-\epsilon_{\theta}^{\text{guided}}(x_t,t,c)\right\|_2^2\right].
\end{equation}
Importantly, we decouple guidance distillation from step reduction~\cite{meng2023distillationguideddiffusionmodels,salimansprogressive}. We train the student to match the teacher's guided predictions without altering the sampling schedule. This formulation accelerates generation by omitting auxiliary evaluations, while crucially preserving the full diffusion trajectory required by our subsequent preference optimization framework.

\subsection{Diffusion-APO: Trajectory-Aware Preference Alignment}
\label{sec32}
Existing diffusion-based DPO methods typically sample timesteps uniformly or via static distributions, introducing a train-inference distribution shift. In practice, advanced samplers (e.g., DDIM, DPM) only traverse a sparse subset of discrete timesteps. Consequently, optimizing over the full continuous range wastes substantial computational bandwidth on timesteps rarely visited during generation. This temporal mismatch dilutes the preference signal along the true generation path, bottlenecking alignment efficiency.

To resolve this, we propose Diffusion-APO, a trajectory-aware algorithm that tightly couples DPO sampling with the inference-time denoising process. Rather than sampling blindly from the global horizon, Diffusion-APO uses discrete inference steps as fixed anchors, sampling within localized temporal neighborhoods. This approach concentrates the optimization objective on regions that dominate generative quality, while maintaining sufficient stochasticity to ensure training stability and prevent overfitting to a rigid sampling grid.

\textbf{Random Shift and Random Offset.} 
Formally, the timestep sampling strategy in Diffusion-APO operates in two stages. First, we construct a set of anchor trajectories by applying a randomized scaling shift $s$ to the base noise schedule. Given a uniformly distributed noise level $\sigma \in [0,1]$, the shifted noise level $\tilde{\sigma}$ is parameterized as:
\begin{equation}
\tilde{\sigma} = \frac{s \cdot \sigma}{1+(s-1) \cdot \sigma},
\end{equation}
where $s$ is drawn from a discrete, predefined set $\mathcal{S}$. The corresponding diffusion timestep is then mapped via $t=\text{round}(\tilde{\sigma}T)$, where $T$ denotes the maximum timestep. This operation yields a discrete sequence of inference-aligned anchors that mimic the structural properties of practical samplers, while injecting controlled macro-level variability. Second, to prevent overfitting to this rigid schedule and encourage localized exploration, we introduce an instance-level stochastic perturbation. Upon selecting an anchor timestep $t$, the final training timestep $t^{'}$ is sampled by adding a zero-mean Gaussian offset: $t^{'}\sim \mathcal{N}(t,\gamma^2)$, then discretized to the nearest valid inference step and clipped to the interval $[0,T]$. By coupling the schedule-level shift with this Gaussian dispersion, Diffusion-APO constructs a dense optimization landscape that remains strictly concentrated around the true generative path, whilst maintaining sufficient coverage of its immediate temporal vicinity.

\textbf{Hybrid High–low Frequency Sliding Window.} 
Directly optimizing isolated diffusion timesteps often destabilizes training due to strong temporal dependencies and severe gradient variance across noise scales, particularly the volatile gradients in low-noise regimes. To address this, Diffusion-APO introduces a hybrid high-low frequency sliding-window gradient accumulation strategy. We define the sliding window gradient as: $\nabla \mathcal{L}_{total}=\sum_{t\in \mathcal{W}}\nabla\mathcal{L}_{APO}(t)$, where $\mathcal{W}$ denotes a set of $K$ timesteps sampled from two distinct regimes: a high-frequency set $\mathcal{W}_{high}=\left\lbrace t|0.8T\leq t \leq T\right\rbrace$ and a low-frequency set $\mathcal{W}_{low}=\left\lbrace t| 0\leq t <0.8T\right\rbrace$. We aggregate $N_H=4$ timesteps from $\mathcal{W}_{high}$ and $N_L=1$ timestep from $\mathcal{W}_{low}$ per accumulation step. This stabilizes the loss landscape by neutralizing erratic spikes. Furthermore, in the context of on-policy methods, this gradient accumulation operation significantly reduces the proportional overhead of the video generation  process within the complete training pipeline.

\begin{figure}[t]
    \centering
    \includegraphics[width=1.0\linewidth]{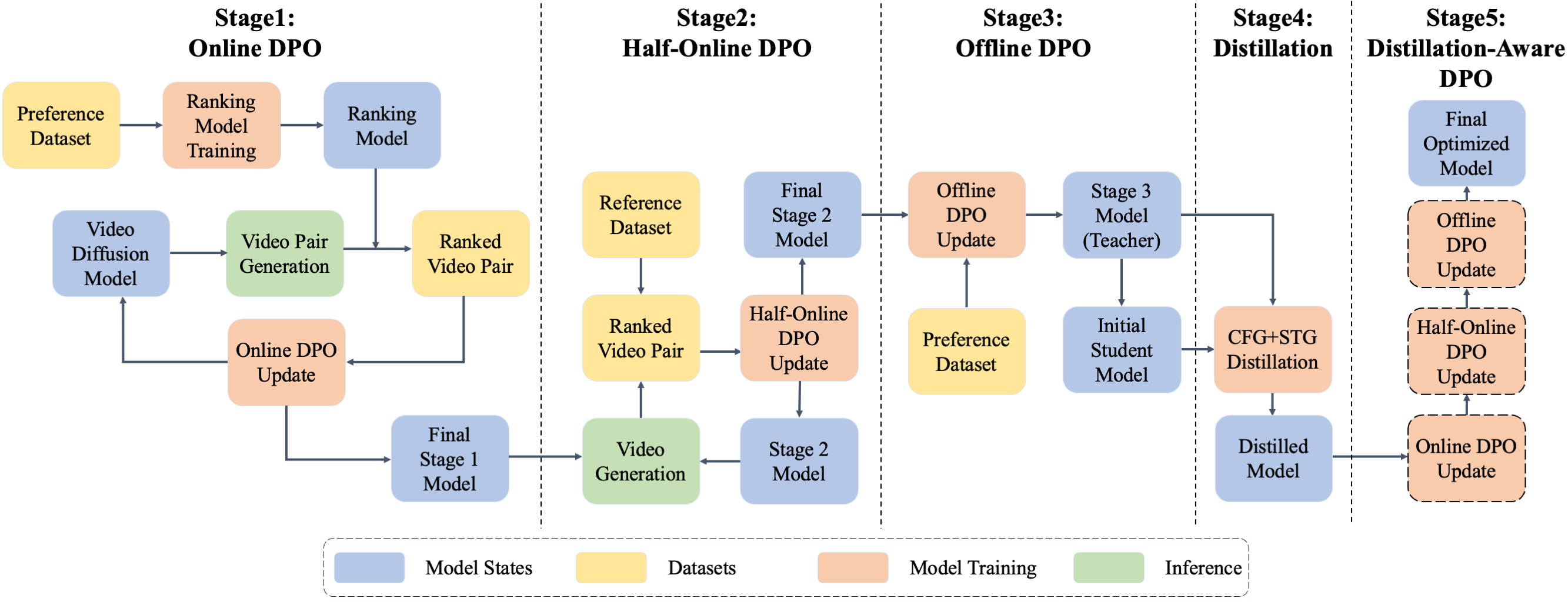}
    \caption{\textbf{Overview of the Unified Preference Optimization Framework.} The five-stage pipeline enables progressive, scalable preference alignment governed by the Diffusion-APO objective.}
    \label{fig:dpo}
\end{figure}

\subsection{Unified Preference Optimization Framework}
\label{sec33}
Beyond the core timestep sampling strategy of Diffusion-APO, we introduce a unified preference optimization framework that synergizes multiple DPO paradigms. This design is driven by two empirical realities: (1) preference data often come from heterogeneous sources with different scales and quality levels, and (2) distinct phases of policy training necessitate tailored alignment strategies. Consequently, our framework seamlessly integrates online, half-online, offline, and distillation-aware DPO into a cohesive progression. This allows the model to dynamically adapt to varying data distributions while monotonically enhancing alignment fidelity, as illustrated in Fig.~\ref{fig:dpo}. For each stage, we initialize the reference policy $\pi_{ref}$ as the model state from the preceding stage and keep it frozen throughout that stage.

The framework operates through a progressive, coarse-to-fine alignment curriculum. Initially, we deploy online and half-online DPO to aggressively scale the preference dataset. By prompting the active policy to generate on-the-fly comparisons, the model extensively explores the generation manifold and bootstraps its own training signal. Following this broad exploratory phase, we transition to standard offline DPO, exploiting high-fidelity, human-annotated preference pairs to rigorously fine-tune the policy and lock in precise alignment. Subsequently, to achieve practical deployment efficiency, we compress the auxiliary signals via guidance distillation (e.g., CFG and STG). Crucially, to counteract any alignment drift incurred during this compression, we conclude with a distillation-aware DPO approach. Throughout this curriculum, every stage is governed by the consistent Diffusion-APO objective. By utilizing this unified goal, our framework provides a robust alternative to reward-based RLHF, ensuring theoretical consistency and algorithmic simplicity across the entire post-training pipeline.

\subsubsection{Online DPO}
To continuously expand the optimization landscape and mitigate distribution shift, Online DPO dynamically bootstraps preference pairs on-the-fly. We utilize a multimodal ranking model, fine-tuned on human annotations, as an automated oracle to assess visual quality and condition adherence. During each training iteration, given an arbitrary input condition (e.g., text prompt and reference image), the active policy generates two candidate videos from distinct noise seeds. These candidates are compared by our ranking model to designate the chosen and rejected samples. This pair is subsequently optimized via the Diffusion-APO objective. This continuous generation-and-ranking cycle enables robust exploration of the generation manifold while scaling preference alignment beyond static datasets.

\subsubsection{Half-Online DPO}
Half-Online DPO bridges the gap between static offline datasets and dynamic exploration by contrasting human-labeled priors with on-policy samples. For a given condition, we retrieve a labeled video from the offline buffer as an anchor and synthesize an on-policy counterpart via the active policy. A positive (chosen) offline anchor implicitly designates our generation as the rejected candidate, and vice versa. This hybrid strategy forces the model to align its generation manifold with human priors, maximizing the utility of sparse manual annotations while stabilizing exploration via the Diffusion-APO objective.

\subsubsection{Distillation-Aware DPO}
While guidance distillation (e.g., CFG/STG absorption) enhances inference efficiency, it often induces alignment drift, as regression-based objectives may overwrite prior human-centric knowledge. To counteract this capacity degradation, we introduce distillation-aware DPO as a terminal fine-tuning phase. By subjecting the distilled student to a final round of Diffusion-APO, we explicitly recover the compromised alignment fidelity. This corrective stage ensures that inference acceleration does not sacrifice generative quality or instruction following.

In summary, our multi-stage curriculum progressively aligns the video diffusion model by seamlessly transitioning across the aforementioned DPO paradigms. Crucially, rather than treating these phases as isolated processes, the entire pipeline is strictly unified under the Diffusion-APO objective. Unlike SPIN~\cite{chen2024self}, our formulation specifically resolves diffusion-inherent train-inference timestep discrepancies, ensuring robust and monotonic improvements in generative fidelity across varying data distributions and computational constraints.

\section{Experiments}

\subsection{General Setup}
\textbf{Implementation Details.} We utilize Wan2.1 and Wan2.2~\cite{wan2025wan} as foundational image-to-video (I2V) models, initially fine-tuned on a large-scale internal dataset to establish a robust baseline. Then we employ LoRA for parameter-efficient fine-tuning for preference alignment, optimized with AdamW (learning rate $=1\times 10^{-4}$) across 192 NVIDIA H200 GPUs. Given computational constraints and the closed-source nature of certain state-of-the-art architectures, our experiments focus on demonstrating the relative training efficiency and alignment upper-bound improvements of Diffusion-APO against standard DPO on these identical baselines. All experiments adhere to a standardized specification: $720\times960$ resolution (3:4 aspect ratio), 24 FPS, and 5-second (121-frame) duration. For Diffusion-APO hyperparameters, the sliding window size is set to 5 (4 high-frequency and 1 low-frequency timestep) with a stride of 4. The schedule shift $s$ is sampled from $\left\lbrace3,4,5,6\right\rbrace$, and the temporal offset is drawn from $\mathcal{N}(0,5)$. The total compute required for the entire five-stage pipeline is estimated at approximately 64,000 GPU-hours on H200 nodes.

\textbf{Evaluation Protocol.} We benchmark Diffusion-APO against standard Diffusion-DPO to demonstrate superior convergence efficiency without scalar rewards, and validate our unified framework’s refinement capabilities. Given that automated benchmark metrics such as VBench~\cite{huang2023vbench,zheng2025vbench2,huang2025vbench++} currently exhibit a saturation effect with constrained dynamic ranges for state-of-the-art models and often struggle with nuanced generative artifacts, we primarily rely on rigorous human evaluation from 24 evaluators across two dimensions: (1) visual quality and (2) instruction following. Our benchmarks include 400 test cases for quality and 256 for instruction following. Subjective robustness is ensured by three independent annotators, with an inter-rater agreement exceeding 90\% (Fleiss’ kappa = 0.82), confirming the consistency and precision of our results. VBench results are added in Appendix~\ref{appendix_vbench}.

\subsection{Diffusion-APO Evaluation}
We curate our alignment dataset by synthesizing 30k I2V pairs, yielding 19k rigorously human-annotated samples. Each pair contrasts a high-fidelity (chosen) video against a distorted (rejected) counterpart to explicitly penalize visual artifacts. Starting from a fine-tuned Wan2.1 baseline, we benchmark Diffusion-APO against standard Diffusion-DPO. To ensure computational parity, DPO is trained for 500 steps, while Diffusion-APO is trained for 100 steps—a configuration that inherently balances our 5-timestep sliding-window accumulation.

\begin{figure}[t]
    \centering
    \includegraphics[width=0.9\linewidth]{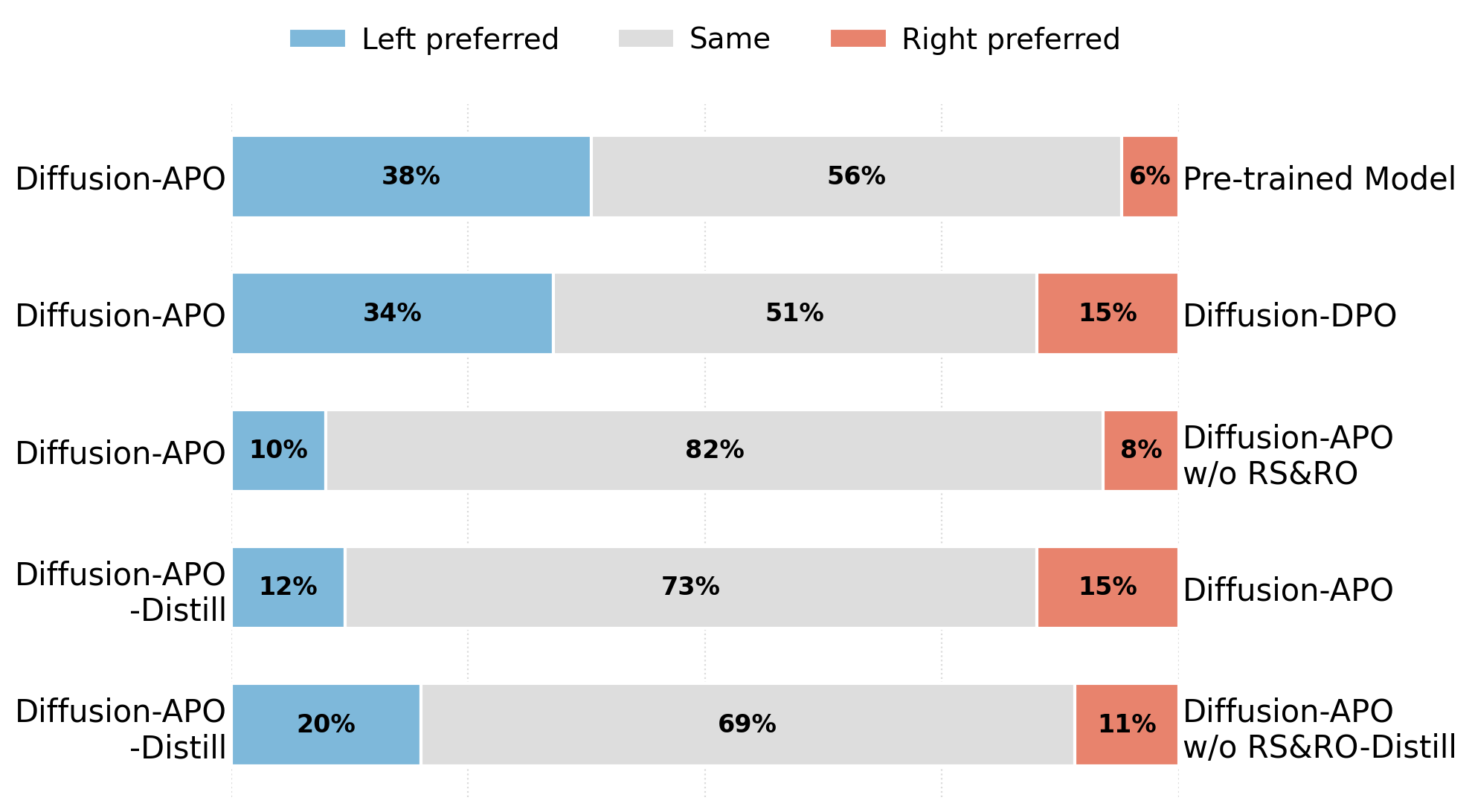}
    \caption{\textbf{Human Preference Evaluation.} Evaluators performed side-by-side, double-blind comparisons of videos generated under identical conditions, selecting the counterpart that demonstrates superior photorealism and reduced visual artifacts across our video quality benchmark (400 videos).}
    \label{fig:win rate}
\end{figure}

Evaluation is conducted on 400 test prompts. Annotators perform two assessments: (1) quantification of absolute distortion rates, and (2) blind pairwise comparisons. As shown in Tab.~\ref{tab:distort1} and Fig.~\ref{fig:win rate}, Diffusion-APO drastically reduces distortion rates and consistently outperforms standard DPO under equivalent computational budgets. Qualitatively, fixed-seed evaluations demonstrate that Diffusion-APO achieves superior temporal coherence—effectively mitigating motion-induced artifacts, preserving object consistency, and generating physically plausible dynamics.

\begin{table}[t]
    \centering
    \caption{\textbf{Distortion Rate Evaluation.} Diffusion-APO consistently achieves lower distortion rates than standard Diffusion-DPO, confirming superior generative quality. All quantitative results (averages and standard deviations) are computed across three distinct random seeds for every test case.}
    \begin{tabular}{l|c|c|c|c}
    \toprule
    Model & \makecell[cc]{Pre-trained\\ Model} & Diffusion-DPO & Diffusion-APO & \makecell[cc]{Diffusion-APO \\ (w/o RS \& RO)}  \\
    \midrule
    Distortion Rate & 21.0\% $\pm$ 2.0\% & 15.0\% $\pm$ 1.5\% & \textbf{8.0\% $\pm$ 1.0\%} & 9.0\% $\pm$ 1.0\%  \\
    \bottomrule
    \end{tabular}
    \label{tab:distort1}
\end{table}

\begin{figure}[t]
    \centering
    \includegraphics[width=1.0\linewidth]{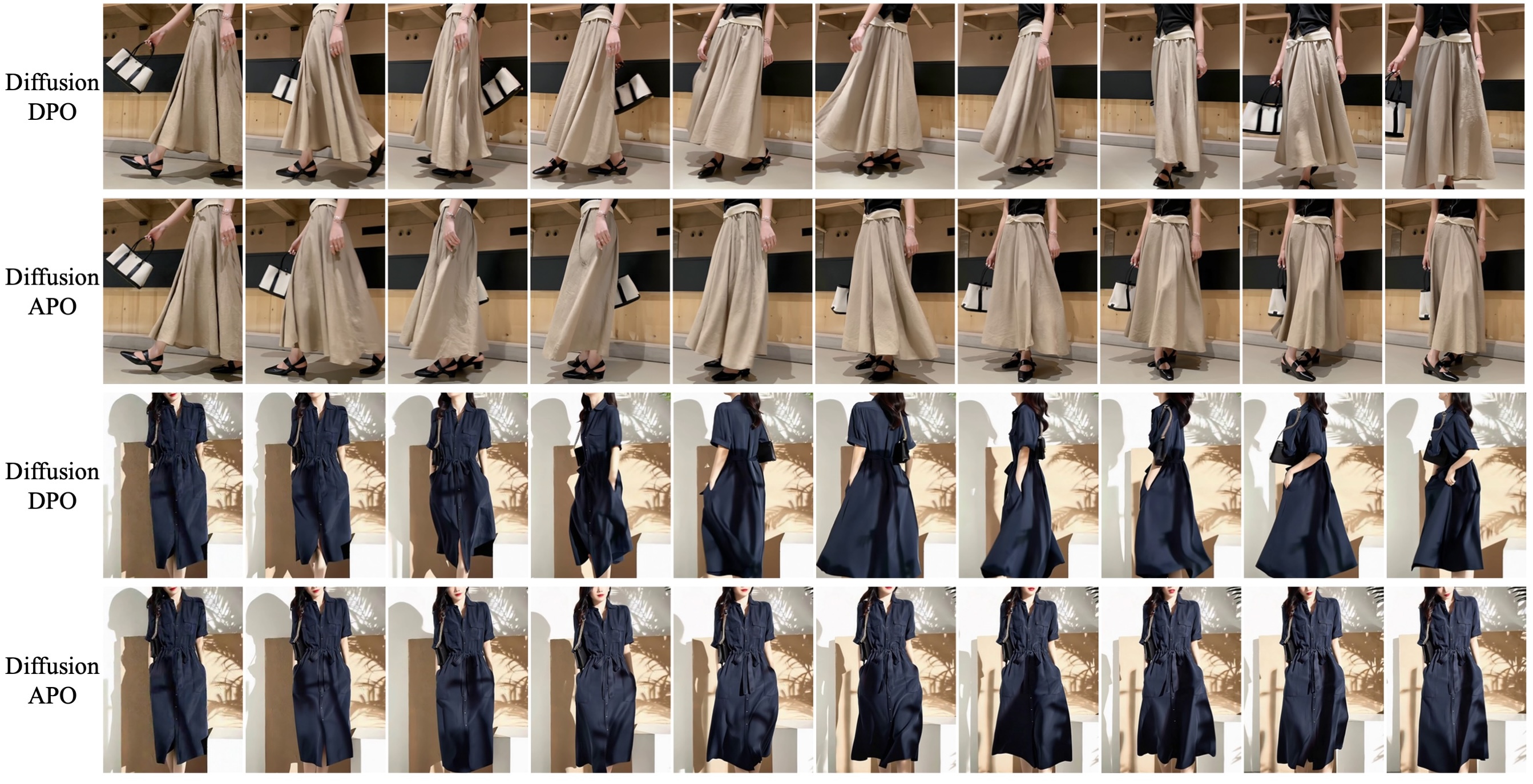}
    \caption{\textbf{Qualitative Comparison: Diffusion-DPO vs. Diffusion-APO.} Under identical training configurations and datasets, Diffusion-APO produces videos with substantially fewer structural artifacts and superior visual fidelity. This performance gap underscores the efficacy of our trajectory-aware, inference-aligned training strategy in optimizing the generative process.}
    \label{fig:vis1}
\end{figure}

\textbf{Ablations on Random Shift (RS) and Random Offset (RO).} We first evaluate the stochastic components of our timestep sampling: random schedule shifts and temporal offsets. As illustrated in Fig.~\ref{fig:win rate}, omitting these perturbations yields only marginal degradations in pre-distillation performance. However, their critical utility emerges post-distillation: policies trained with these operations exhibit remarkable resilience to capacity degradation, maintaining consistently lower distortion rates. This confirms that while localized stochasticity is secondary for initial alignment, it injects vital structural robustness into the policy, effectively buffering against alignment drift during model compression.

\textbf{Ablations on Hybrid High-Low Frequency Sliding Window.} We further evaluate the hybrid temporal sliding-window mechanism. Removing this prior exposes the optimization landscape to severe loss oscillations across varying noise scales. Specifically, isolating low-frequency temporal regimes triggers a sharp spike in visual artifacts; conversely, relying exclusively on high-frequency regimes leads to suboptimal convergence. By ensembling high and low-frequency steps within a single sliding window, our strategy acts as a crucial variance-reduction mechanism, neutralizing performance oscillations and facilitating the convergence of a stable, artifact-free policy. We provide a quantitative loss comparison in Fig.~\ref{fig:loss_compare} of Appendix~\ref{appendix_loss_curve}, which illustrates how our strategy eliminates the erratic loss spikes observed in baseline configurations.

\begin{table}[t]
    \centering
    \caption{\textbf{Quantitative human evaluation across optimization stages.} Configurations are defined as: \textbf{Wan2.2} (SFT baseline); \textbf{Baseline} (Wan2.2 + SFT); \textbf{Ours (APO)} (Baseline + Diffusion-APO); \textbf{Distilled} (Ours (APO) + Guidance Distillation); and \textbf{Ours (Distilled)} (Ours (APO) + Guidance Distillation + Distillation-Aware DPO). Evaluations are conducted on 656 test samples via 3-point mean opinion score (0-2) across five dimensions. Best results are \textbf{bold}.}
    \begin{tabular}{l|c|c|c|c|c|c}
    \toprule
    Models & \makecell[c]{Visual \\ Plausibility} & \makecell[c]{Human \\Plausibility} & \makecell[c]{Subject\\ Consistency} & \makecell[c]{Motion \\Aesthetics} & \makecell[c]{Instruction \\Following} & \makecell[c]{Overall \\Scores} \\ \midrule
    Wan2.2 & 1.64 & 1.16 & 1.24 & 1.67 & 1.61 & 7.32 \\
    Baseline & 1.74 &1.35 & 1.22 &1.79 &1.54 & 7.64 \\
    Ours (APO) & 1.91 &\bf{1.41} &1.30 & \bf{1.85} &1.77 & 8.12 \\
    Distilled & 1.84 & 1.37 & 1.28 & 1.81 & 1.76 & 8.06 \\
    Ours (Distilled) & \bf{1.92} &1.39 &\bf{1.32} & \bf{1.85} &\bf{1.78} & \bf{8.26} \\
    \bottomrule
    \end{tabular}
    \label{tab:overall_eval}
\end{table}

\subsection{Unified Preference Optimization Framework Validation}
We systematically validate our framework using the fine-tuned Wan2.2 baseline. Our stage-wise design isolates the contribution of each alignment paradigm: starting from the foundational model, we first integrate Diffusion-APO to achieve robust preference alignment, followed by model distillation and a final distillation-aware fine-tuning phase. Throughout this progression, the Diffusion-APO objective remains the consistent core, ensuring theoretical unity and algorithmic simplicity across all stages of the optimization pipeline.

\textbf{Online DPO: Scalable Artifact Reduction.} We first deploy Online DPO to scale preference alignment for visual fidelity. To bootstrap the preference signal, we fine-tune a Qwen3-VL-8B model on human-annotated preference pairs, serving as a pairwise ranker that explicitly penalizes geometric distortions and prioritizes photorealism. During training, the active diffusion model dynamically synthesizes candidate pairs from diverse prompts and images, which are immediately ranked by the vision-language model (VLM) to form preference data. After 300 iterations, we observe a substantial improvement in visual fidelity, with the overall distortion rate dropping from 14.0\% to 7.5\%.

\textbf{Half-Online DPO: Enhancing Instruction Following.} With a stabilized visual baseline, we employ Half-Online DPO to enforce fine-grained instruction following. Our offline dataset comprises 30k annotated anchors: 18k positive pairs with high text-video alignment and no deformities, and 12k mismatched negative anchors. We query the policy to synthesize on-policy counterparts, forming hybrid pairs for training. After 300 iterations, blind human evaluation on a dedicated benchmark shows that our updated policy achieves a win/tie/loss ratio of 86/148/22 against the pre-Half-Online model. Crucially, this improvement in semantic alignment preserves the previously acquired visual fidelity, with the distortion rate remaining stable at 7.5\%. This confirms that Half-Online DPO effectively integrates instructional priors without inducing structural degradation.

\textbf{Offline DPO: Rigorous Human-Driven Refinement.} To conclude the pre-distillation pipeline, we perform Offline DPO on a curated set of 50k human-annotated preference pairs. Serving as the terminal refinement stage, this phase leverages high-fidelity human supervision to rectify geometric artifacts and nuanced quality degradations that often elude automated VLM rankers. Following 300 iterations, the distortion rate is further reduced from 7.5\% to 5.0\%. This result unequivocally validates the critical necessity of high-quality human priors for achieving robust, macroscopic alignment before model compression.

\textbf{Distillation-Aware DPO: Recovering Generative Fidelity.} To mitigate the capacity degradation inherent to model compression, we conclude with a distillation-aware preference optimization phase. We first perform guidance distillation by absorbing both CFG and STG into the base network, yielding a 
$3\times$ inference speedup. While instruction following remains robust post-distillation, we observe a slight regression in visual fidelity, with the distortion rate rising from 5.0\% to 7.0\%. To counteract this, we perform an additional 200 steps of Offline DPO specifically targeted at rectifying these compression-induced artifacts. This corrective phase not only fully restores the generative model's structural integrity but further reduces the distortion rate to 4.5\%, ultimately surpassing the pre-distillation performance.

We summarize the end-to-end efficacy of our Unified Preference Optimization Framework in Tab.~\ref{tab:overall_eval}. The progression from the SFT baseline to our fully optimized \textbf{Ours (Distilled)} configuration demonstrates consistent, monotonic improvements across all evaluated dimensions. While the \textbf{Baseline} establishes a solid generative foundation, the integration of our core Diffusion-APO objective (\textbf{Ours (APO)}) triggers a substantial performance leap, driving the overall score from 7.64 to 8.12, with notable gains in Human Plausibility (1.41) and Motion Aesthetics (1.85). Most impressively, our final \textbf{Ours (Distilled)} model—which incorporates guidance distillation followed by distillation-aware DPO—successfully mitigates the quality degradation typically associated with model compression, achieving the highest Overall Score (8.26). By peaking in Subject Consistency (1.32) and Instruction Following (1.78), \textbf{Ours (Distilled)} conclusively validates that our unified pipeline delivers a highly accelerated inference model while enhancing generative fidelity. These stage-wise gains reinforce our core claim: a multi-paradigm approach governed by the Diffusion-APO objective is essential for unlocking the generative potential of DiT architectures without the computational overhead of auxiliary reward models. Supplementary quantitative (VBench~\cite{zheng2025vbench2,huang2023vbench,huang2025vbench++}) results and more qualitative comparison are provided in Appendix~\ref{appendix_vbench} and \ref{appendix_experiment}.

\section{Conclusion}
In this work, we introduce Diffusion-APO, which resolves the train-inference discrepancy inherent in standard DPO methods. By anchoring timestep sampling to the discrete inference trajectory and employing a hybrid high-low frequency sliding-window accumulation strategy, Diffusion-APO stabilizes the optimization landscape and concentrates preference gradients precisely where they dictate generation quality. Building upon this, we integrate our objective into a holistic, multi-stage pipeline encompassing online exploration, hybrid half-online anchoring, rigorous offline refinement, and distillation-aware drift rectification. This coarse-to-fine unified preference optimization framework provides a flexible, end-to-end solution for aligning large-scale video diffusion models with complex human priors, simultaneously advancing training efficiency and deployment scalability. We believe this work establishes a robust foundation for future research in scalable and artifact-free generative video alignment. Limitations of our approach are discussed in Appendix~\ref{appendix_limitations}.

{
    \small
    \bibliographystyle{abbrv}
    \bibliography{main}
}

\clearpage
\appendix

\section{Limitations}
\label{appendix_limitations}
Although Diffusion-APO effectively mitigates train-inference timestep discrepancies, its optimization efficacy remains fundamentally bottlenecked by the fidelity of the preference signals. Specifically, our reliance on ranking signals from automated vision-language models (VLMs) or human annotators can occasionally result in noisy targets, particularly when failing to capture highly nuanced temporal inconsistencies or imperceptible structural artifacts. Furthermore, while our unified multi-stage RLHF pipeline—comprising online, half-online, offline, and distillation-aware stages—progressively resolves distinct alignment challenges, this sequential curriculum increases training complexity and cumulative computational overhead, which may challenge its deployment in resource-constrained settings. Finally, as our empirical validation is centered on specific foundational I2V architectures, generalizing this framework across a broader spectrum of video diffusion models remains a critical avenue for future research.

\section{Broader Impact}
\label{appendix_impact}
This framework enhances the visual fidelity and controllability of video diffusion models, offering significant potential for applications in content creation, virtual production, and human–computer interaction. By effectively mitigating structural artifacts and improving instruction adherence, our method enables the generation of high-quality, physically plausible, and semantically aligned media.

However, we acknowledge that advancements in high-fidelity video generation may be misused to create misleading or deceptive content, which poses risks to information integrity and public trust. To mitigate these concerns, we emphasize that technical improvements must be accompanied by robust safeguards and responsible deployment practices. We advocate for continued research into synthetic media detection, provenance tracking, and content governance to ensure that these generative tools are used ethically and safely. We are committed to fostering a research environment that balances generative capability with accountability.

\section{Specialized Ranking Model for Online DPO}
\label{appendix_ranker}

In the Online DPO stage, providing fine-grained and accurate preference signals is critical for effective alignment. Given that the evaluation criteria for video generation vary significantly across different content domains, relying on a single monolithic ranking model often leads to suboptimal results. Specifically, human-centric videos are highly susceptible to anatomical and limb distortions, whereas non-human-centric videos (e.g., landscapes, objects) are more prone to physical law violations and illogical camera movements. To address this discrepancy, we decouple the ranking modeling process and train two domain-specialized ranking models.

\subsection{Data Collection and Negative Mining}

\textbf{Data Collection.} 
During the initial data collection phase, human annotators were instructed to strictly focus on geometric distortion and visual realism (e.g., naturalness of motion). We initially collected and human-annotated 57k human-centric video pairs and 15k non-human-centric video pairs.

\textbf{Negative Mining.} 
A common failure mode in video generation models is the \textbf{static bias}, generating near-static videos to safely avoid temporal inconsistencies and geometric deformations. To break this generative shortcut and enforce true dynamic alignment, we artificially construct slow-motion videos using frame dropping and temporal interpolation algorithms. When paired with standard videos of identical single-frame quality, this design forces the ranking model to explicitly evaluate ``motion rhythm'' and ``dynamic amplitude'', thereby significantly enhancing its perception of dynamic coherence. This strategy contributes an additional 7k human-centric and 3k non-human-centric samples. Ultimately, the training dataset comprises 64k pairs for the human-centric ranking model and 18k pairs for the non-human-centric ranking model.

\subsection{Model Architecture and Training Design}

We select \textbf{Qwen3-VL-8B} as the foundational backbone for our ranking models. To ensure the model adequately captures long-term temporal dependencies, the input sequence for both training and inference is uniformly standardized to $40$ frames. Furthermore, to prevent the vision-language model from confusing the spatial and temporal order of the two compared videos, we explicitly inject \texttt{vision\_id} tags at the input stage.

Traditional ranking models typically output a simple binary choice (e.g., A or B), which lacks interpretability and dense guidance. To overcome this, we design a \textbf{Hierarchical Evaluation Prompt}. Instead of a mere partial-order judgment, the model is required to output a 1-out-of-4 explanatory label. For instance, the model must differentiate whether a video is superior due to ``fewer geometric distortions'' or because ``distortions are comparable, but it exhibits richer dynamic motion.'' This design substantially improves both label interpretability and the learning efficiency of the model. Below, we detail the exact prompt templates used to instruct the Vision-Language Model for hierarchical preference ranking.

\paragraph{Prompt for Human-Centric Videos}
\begin{quote}
\small
\texttt{<video><video>} \\
You are sequentially given two videos, Video 1 and Video 2. Please evaluate them based on the following criteria and select the better video:

\textbf{1. Primary Criterion - Degree of Visual Distortion:}
\begin{itemize}
    \item \textbf{a) Severe Defects:} Anomalies violating physical laws (e.g., anti-gravity cars, unnatural movement of objects without applied force), unnatural transitions/black screens/objects appearing or disappearing out of nowhere, severe anatomical distortions (severed limbs, misplaced facial features), or visible mosaic/glitches in the frame.
    \item \textbf{b) Obvious Issues:} Clothing/accessories dissolving or misaligned (e.g., disappearing earrings), abnormal limb swapping (e.g., left becoming right), confirmed finger deformities (e.g., extra fingers/webbed fingers).
    \item \textbf{c) Minor Flaws:} Slight single-frame hand deformation (5 fingers present but abnormally bent), minor clipping of hair/clothing (area $< 5\%$), transient errors in environmental textures.
    \item \textbf{d) Qualified Frame:} Physically accurate motion blur, pixel-level issues invisible to the naked eye.
\end{itemize}

\textbf{2. Secondary Criterion - Dynamic Performance:}
\begin{itemize}
    \item When both videos have no obvious distortions, prioritize the video with larger motion amplitude and a more vibrant visual dynamic.
    \item Avoid selecting completely static or nearly motionless videos.
\end{itemize}

Please compare Video 1 and Video 2, and output \textbf{ONLY} one of the following four options:
\begin{enumerate}
    \item Video 1 has a lower degree of visual distortion.
    \item Video 2 has a lower degree of visual distortion.
    \item Both videos have similar degrees of distortion, but Video 1 features larger camera movement or richer human actions.
    \item Both videos have similar degrees of distortion, but Video 2 features larger camera movement or richer human actions.
\end{enumerate}
\end{quote}

\paragraph{Prompt for Non-Human-Centric Videos}
\begin{quote}
\small
\texttt{<video><video>} \\
You are sequentially given two videos, Video 1 and Video 2. Please strictly evaluate them based on the following criteria and select the better video:

Evaluate whether the following issues exist in the videos (ordered by severity):

\textbf{1. Severe Visual Anomalies (Any occurrence leads to an immediate low-quality rejection):}
\begin{itemize}
    \item Motions violating physical laws (unnatural movement/jittering/floating of objects without applied forces).
    \item Unnatural twisting, deformation, merging, or blurring of objects.
    \item Appearance of mosaics or digital glitches.
    \item Sudden visual jumps, PPT-like fading, or discontinuous transitions.
    \item Objects or textures appearing/disappearing/flickering out of nowhere.
    \item Abnormal changes in object proportions as the camera moves.
    \item Splitting, repeating, or overlapping of objects, textures, or patterns.
    \item Garbled or unrecognizable generated text.
\end{itemize}

\textbf{2. Obvious Visual Issues:}
\begin{itemize}
    \item Abnormal surface textures (suddenly floating, warping, or distorting).
    \item Sudden appearance of mosaics in the frame.
    \item Unnatural material representation (e.g., sudden changes in leather or metal textures).
    \item Unnatural blurring or over-sharpening of object edges.
    \item Illogical dynamic changes in patterns, motifs, or text.
    \item Shadows and lighting that defy physical laws.
    \item Local or global blurring/ghosting.
\end{itemize}

\textbf{3. Visual Coherence Issues:}
\begin{itemize}
    \item Discontinuous object motion trajectories.
    \item Unnatural scene transitions.
    \item Camera shaking or jittering.
\end{itemize}

\textbf{4. Dynamic Performance Evaluation:}
\begin{itemize}
    \item Visual changes should be natural and smooth; avoid selecting essentially static videos.
    \item When visual quality is comparable, prioritize the video with natural camera movements and dynamic visual changes.
\end{itemize}

Please strictly inspect each video for the above issues, paying primary attention to Severe Visual Anomalies. If any severe visual anomaly occurs, immediately classify that video as the inferior option.

Please compare Video 1 and Video 2, and output \textbf{ONLY} one of the following four options:
\begin{enumerate}
    \item Video 1 has a lower degree of visual distortion/anomalies.
    \item Video 2 has a lower degree of visual distortion/anomalies.
    \item Both videos have similar degrees of distortion, but Video 1 features larger/better camera movement.
    \item Both videos have similar degrees of distortion, but Video 2 features larger/better camera movement.
\end{enumerate}
\end{quote}

\subsection{Evaluation Performance}

Evaluated on strictly aligned internal human-preference test sets, our specialized ranking models demonstrate exceptional precision and recall. The \textbf{Human-Centric Ranking Model} achieves accuracy/recall rates of $91.8\%$ and $94.2\%$ across two distinct validation sets, proving highly capable of capturing extremely subtle limb and motion distortions. Meanwhile, the \textbf{Non-Human-Centric Ranking Model}, when evaluated on complex camera movements and open-domain physical scenes, achieves an accuracy of $78.5\%$. These robust metrics ensure that the ranking models provide highly reliable feedback signals for the subsequent RLHF pipeline.

\section{Supplemental Quantitative Results of VBench}
\label{appendix_vbench}
To complement our human evaluations, we provide a quantitative analysis using the VBench-I2V benchmark~\cite{huang2023vbench,huang2025vbench++,zheng2025vbench2}. As shown in Tab.~\ref{tab:vbench-i2v}, we benchmark the foundational Wan2.2-I2V-A14B against our final model derived from the complete Unified Preference Optimization Framework. While VBench provides a useful objective reference, the resulting scores often exhibit a constrained dynamic range. This saturation effect, prevalent in state-of-the-art generative models, suggests that marginal improvements in automated metrics frequently mask substantial gains in perceived visual fidelity, temporal coherence, and instruction adherence—nuances more accurately captured by our rigorous human evaluation protocol. As shown in Tab.~\ref{tab:vbench-i2v}, while the baseline exhibits strong performance in some metrics, our method still achieves superior dynamic degree and imaging quality, indicating that our framework prioritizes active synthesis and motion diversity.

In summary, while we acknowledge the limitations of automated benchmarks in capturing the full nuance of video quality, these VBench results serve as a robust, complementary indicator that our unified framework successfully enhances the model's foundational generative quality.

\begin{table}[htbp]
    \centering
    \small
    \begin{tabular}{c|c|c|c|c|c|c}
    \toprule
     Model & \makecell[c]{subject \\ consistency} & \makecell[c]{background \\ consistency} & \makecell[c]{motion\\smoothness} & \makecell[c]{dynamic \\degree} & \makecell[c]{aesthetic \\ quality} & \makecell[c]{imaging \\quality} \\ \midrule
     \makecell[c]{Wan2.2-I2V-A14B \\ (aspect 16:9) } & 94.68\% & 96.72\% & 98.32\% & 52.85\% & 61.53\% & 70.36\% \\ \hline
     Ours (aspect ratio 16:9) & 95.40\% & 96.80\% & 98.87\% & \textbf{57.50\%} & 63.14\% & 71.81\% \\ \hline
     Ours (aspect ratio 1:1) & \textbf{95.69\%} & \textbf{97.19\%} & \textbf{98.90\%} & 54.50\% & \textbf{64.00\%} & \textbf{72.02\%} \\ \hline
     Ours (aspect ratio 7:4) & 95.35\% & 96.85\% & 98.89\% & 56.93\% & 63.35\% & 71.92\% \\ \hline
     Ours (aspect ratio 8:5) & 95.47\% & 96.86\% & 98.89\% & 54.69\% & 63.39\% & 71.96\% \\ 
    \bottomrule
    \end{tabular}
    \caption{Evaluation of VBench-I2V~\cite{huang2023vbench,huang2025vbench++,zheng2025vbench2} benchmark results. We compare our framework against the baseline Wan2.2-I2V-A14B under various aspect ratios. Results demonstrate that our method consistently maintains or improves performance across key metrics such as subject consistency, motion smoothness, dynamic degree, and imaging quality.}
    \label{tab:vbench-i2v}
\end{table}

\section{Training Stability of the Hybrid High-Low Frequency Sliding Window}
\label{appendix_loss_curve}
\begin{figure}[h]
    \centering
    \includegraphics[width=1.0\linewidth]{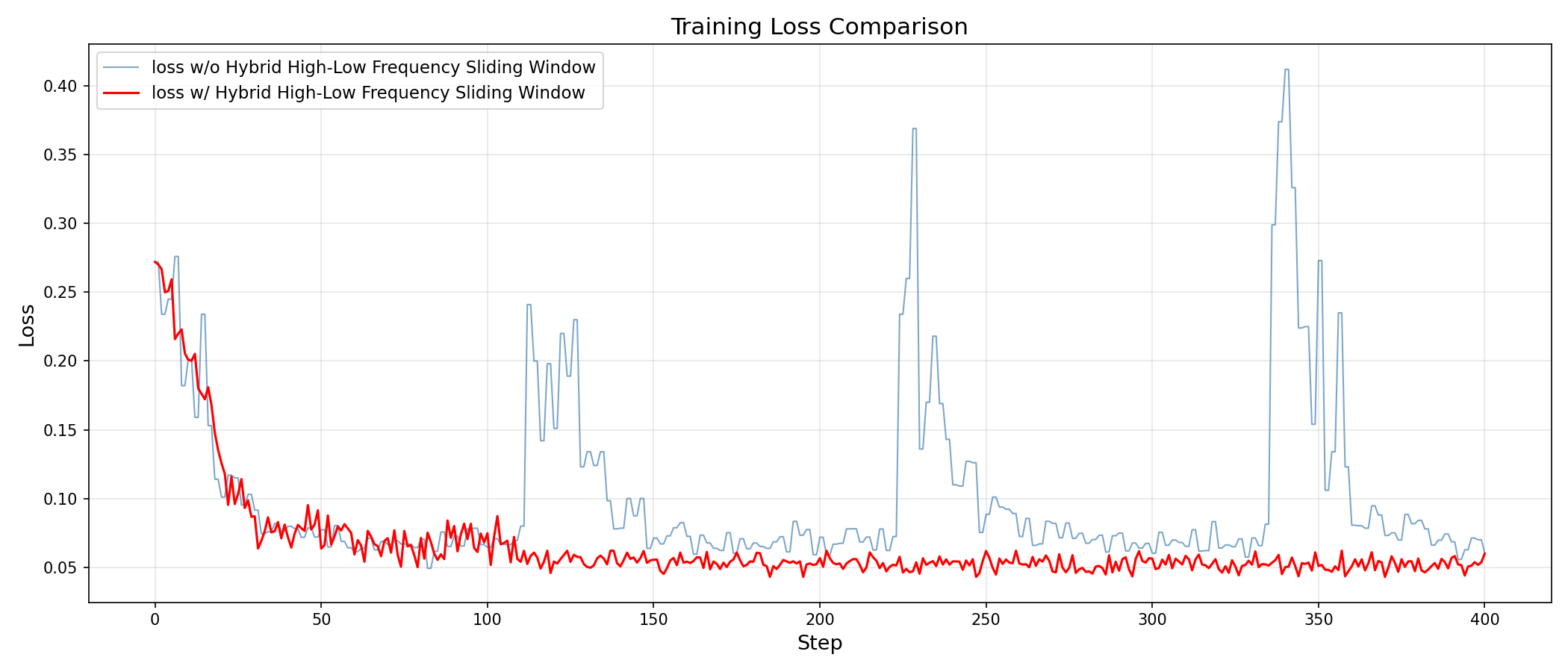}
    \caption{\textbf{Training loss comparison.} The blue line represents the baseline training without the sliding-window mechanism, exhibiting significant instability and periodic loss spikes due to gradient variance across noise scales. The red line, utilizing our hybrid high-low frequency sliding-window, demonstrates a significantly more stable convergence path, which directly translates to improved visual coherence and reduced structural artifacts in the generated videos.}
    \label{fig:loss_compare}
\end{figure}

To substantiate our claim regarding the stability of the Hybrid High-Low Frequency Sliding Window mechanism, we visualize the training loss trajectories in Fig.~\ref{fig:loss_compare}. Optimizing isolated diffusion timesteps (blue line) leads to severe loss oscillations and erratic spikes, particularly in low-noise regimes where gradient variance is high. In contrast, our sliding-window accumulation (red line) acts as a variance-reduction mechanism, smoothing the loss landscape and ensuring stable convergence. This empirical stability is critical for preventing the model from overfitting to localized artifacts that emerge during the erratic spikes of un-smoothed training.

\section{Progressive Visual Validation of the Unified Framework}
\label{appendix_experiment}

To complement the quantitative results presented in the main text, this appendix provides qualitative and empirical evidence illustrating the evolutionary trajectory and training stability of our framework. To ensure a rigorous comparison, we utilize identical random seeds for all side-by-side visual validations. \textbf{All video examples referenced are provided in the supplementary material.}

\subsection{Visual Validation of the Unified Framework}

\begin{figure}[t]
    \centering
    \includegraphics[width=1.0\linewidth]{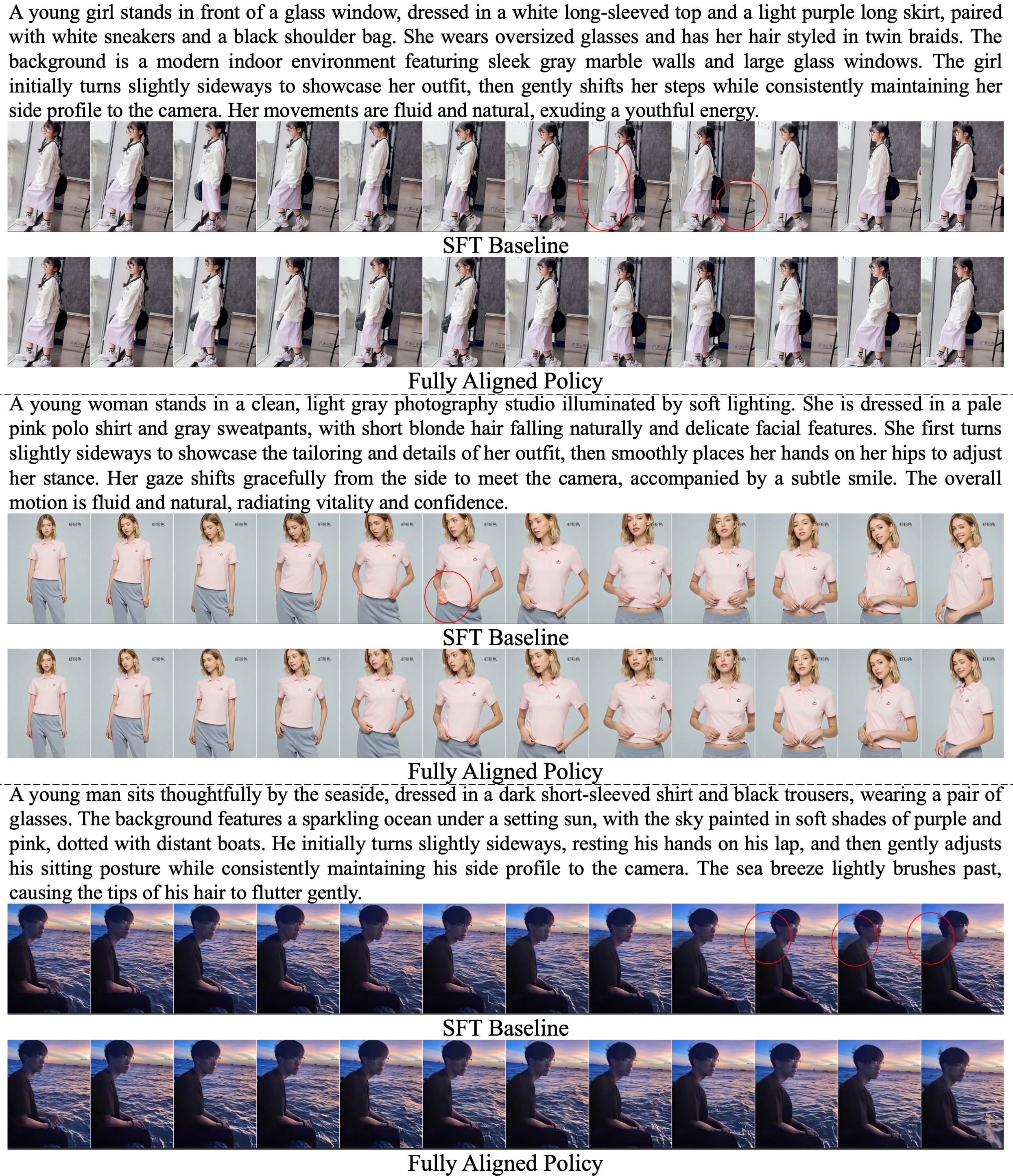}
    \caption{\textbf{Qualitative comparison of generated results.} The SFT baseline model (upper row) suffers from common generative failures in complex scenarios, including severe background distortion, motion-induced deformation, spurious artifacts (circled in red). In contrast, the model fine-tuned with the proposed fully aligned policy (bottom) effectively resolves these artifacts, consistently producing temporally coherent, anatomically accurate, and visually realistic videos.}
    \label{app_fig:vis1}
\end{figure}

\begin{figure}[t]
    \centering
    \includegraphics[width=1.0\linewidth]{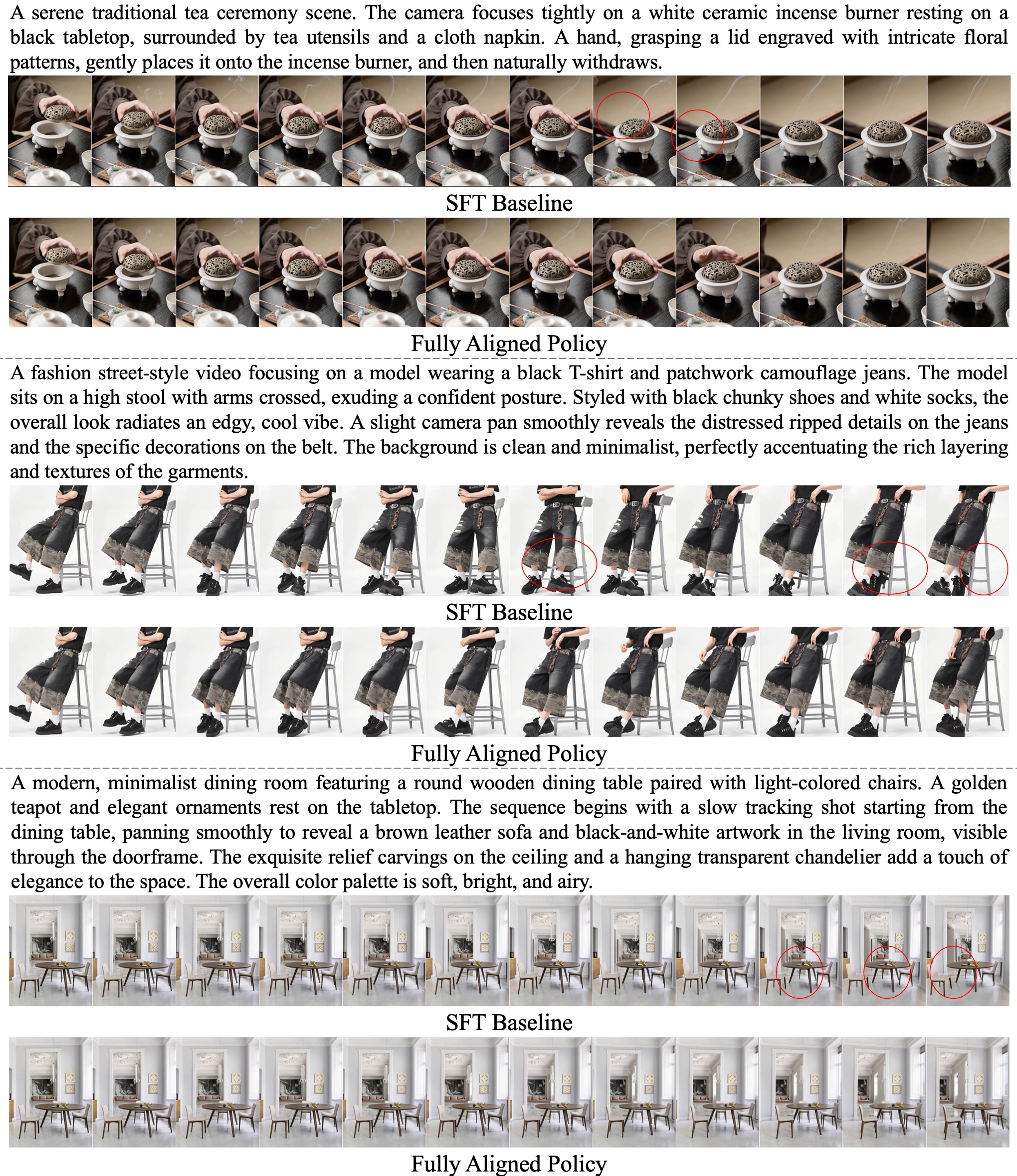}
    \caption{\textbf{Qualitative comparison of generated results.} The SFT baseline model (upper row) suffers from common generative failures in complex scenarios, including severe missing limbs and object deformation (circled in red). In contrast, the model fine-tuned with the proposed fully aligned policy (bottom row) effectively resolves these artifacts, consistently producing temporally coherent, anatomically accurate, and visually realistic videos.}
    \label{app_fig:vis2}
\end{figure}

\begin{figure}[t]
    \centering
    \includegraphics[width=1.0\linewidth]{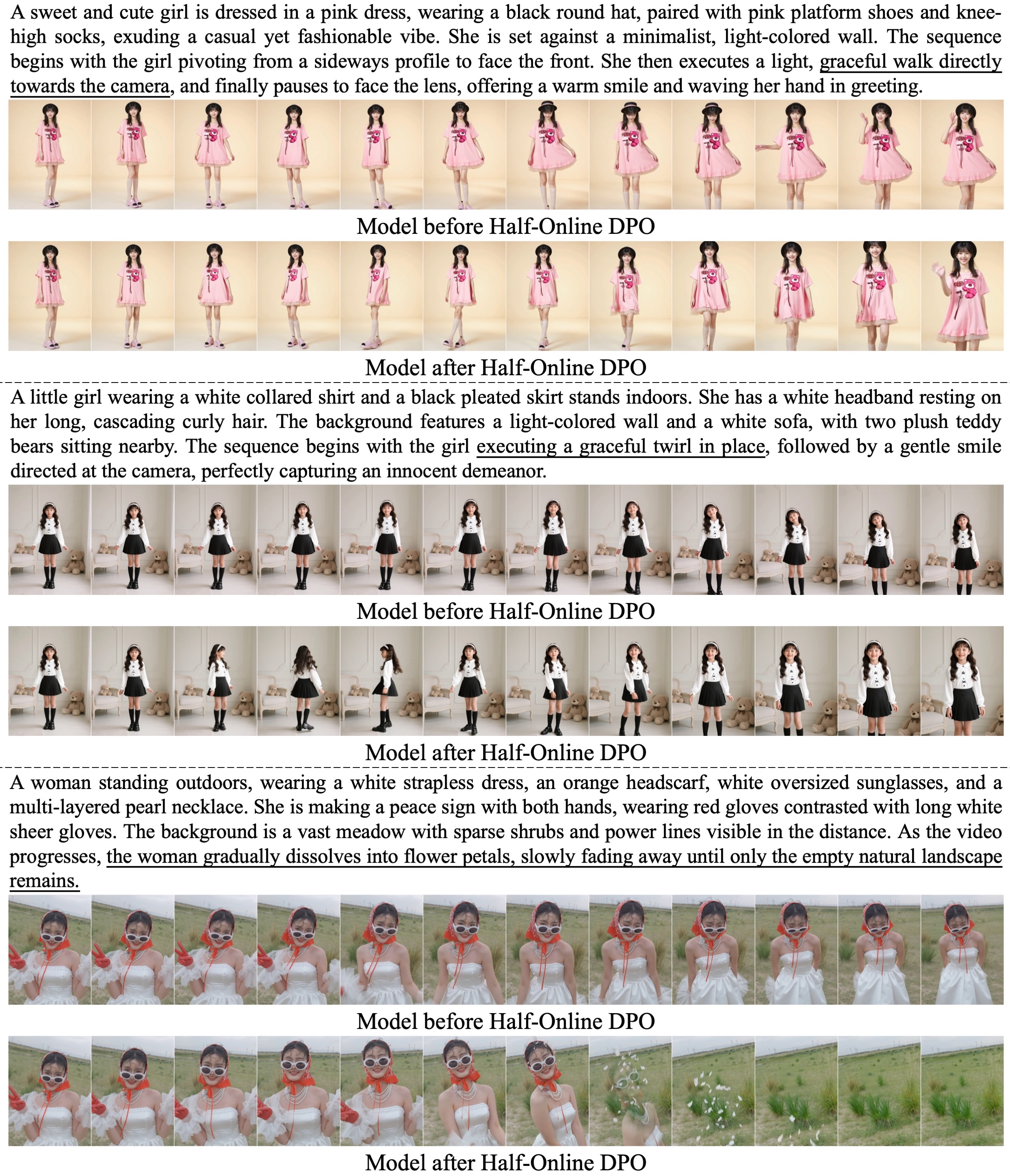}
    \caption{\textbf{Visual demonstration of enhanced instruction following via Half-Online DPO.} In I2V generation settings, the Half-Online DPO stage explicitly improves the model's ability to accurately interpret and execute complex, action-driven text prompts, ensuring the generated visual dynamics perfectly align with user instructions.}
    \label{app_fig:vis4}
\end{figure}

(1) \textbf{Pre-Distillation Alignment}: We contrast the foundational SFT baseline against our policy optimized via the sequential curriculum of Online, Half-Online, and Offline DPO. As demonstrated in Figs.~\ref{app_fig:vis1} and~\ref{app_fig:vis2}, our framework effectively mitigates common generative artifacts—such as background distortions, motion-induced deformation, and anatomical anomalies. While the SFT baseline often exhibits significant structural instabilities (highlighted in red), Diffusion-APO consistently preserves anatomical accuracy and temporal integrity. Furthermore, Fig.~\ref{app_fig:vis4} isolates the impact of the Half-Online DPO stage, confirming its superior capability in achieving semantic alignment with dynamic action prompts during I2V generation.

\begin{figure}
    \centering
    \includegraphics[width=1.0\linewidth]{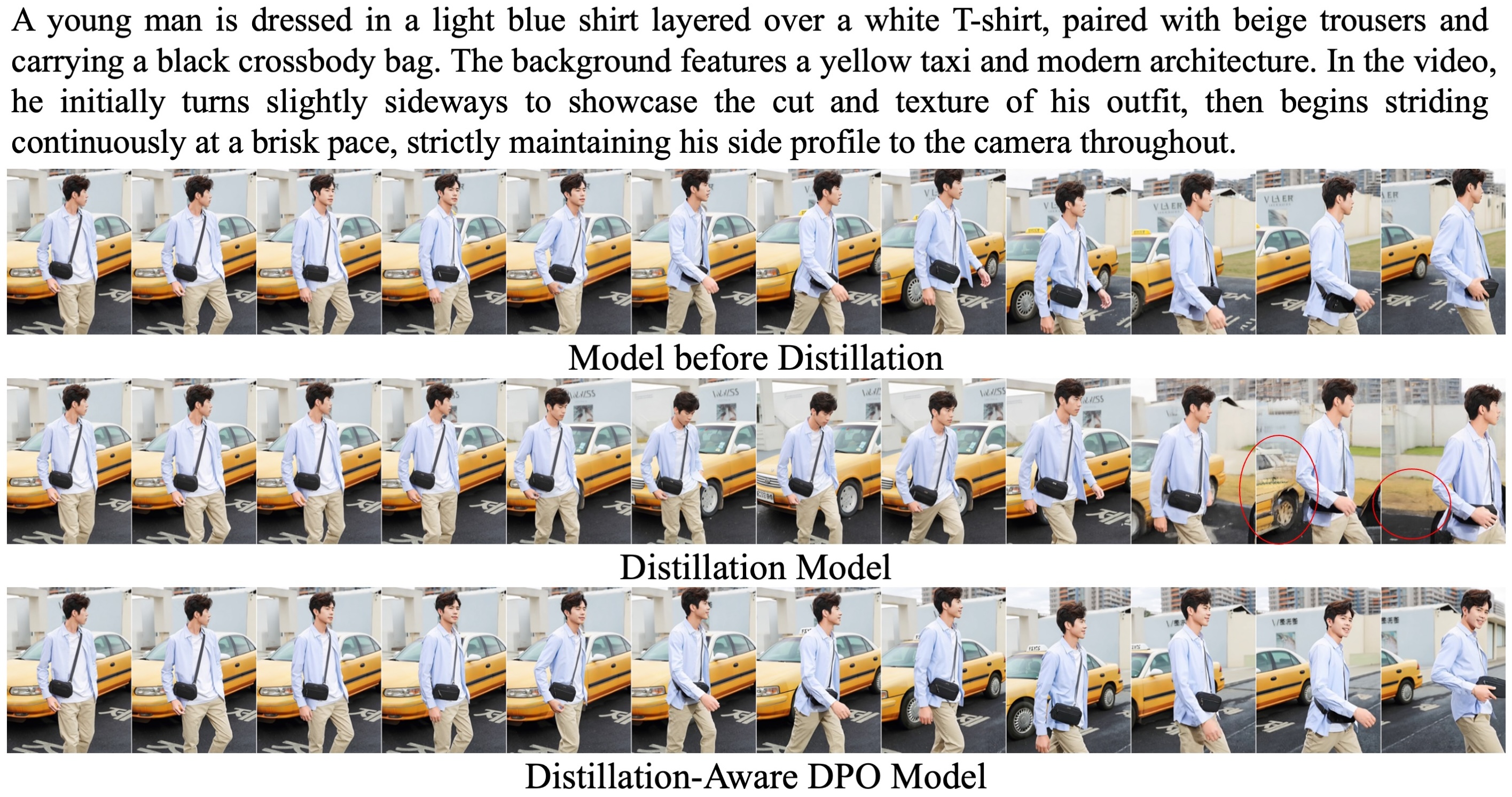}
    \caption{\textbf{Illustrative Example 1 of Post-Distillation Recovery and Enhancement through DPO.} The top row shows the original teacher model's high-fidelity output. The middle row presents the naively distilled student model, which exhibits visual degradation despite a $3 \times$ speedup. The bottom row displays our DPO-optimized model, successfully restoring and often surpassing the teacher's generative fidelity, thus proving DPO's ability to overcome distillation losses and enhance final output while maintaining efficiency.}
    \label{app_fig:vis31}
\end{figure}

\begin{figure}
    \centering
    \includegraphics[width=1.0\linewidth]{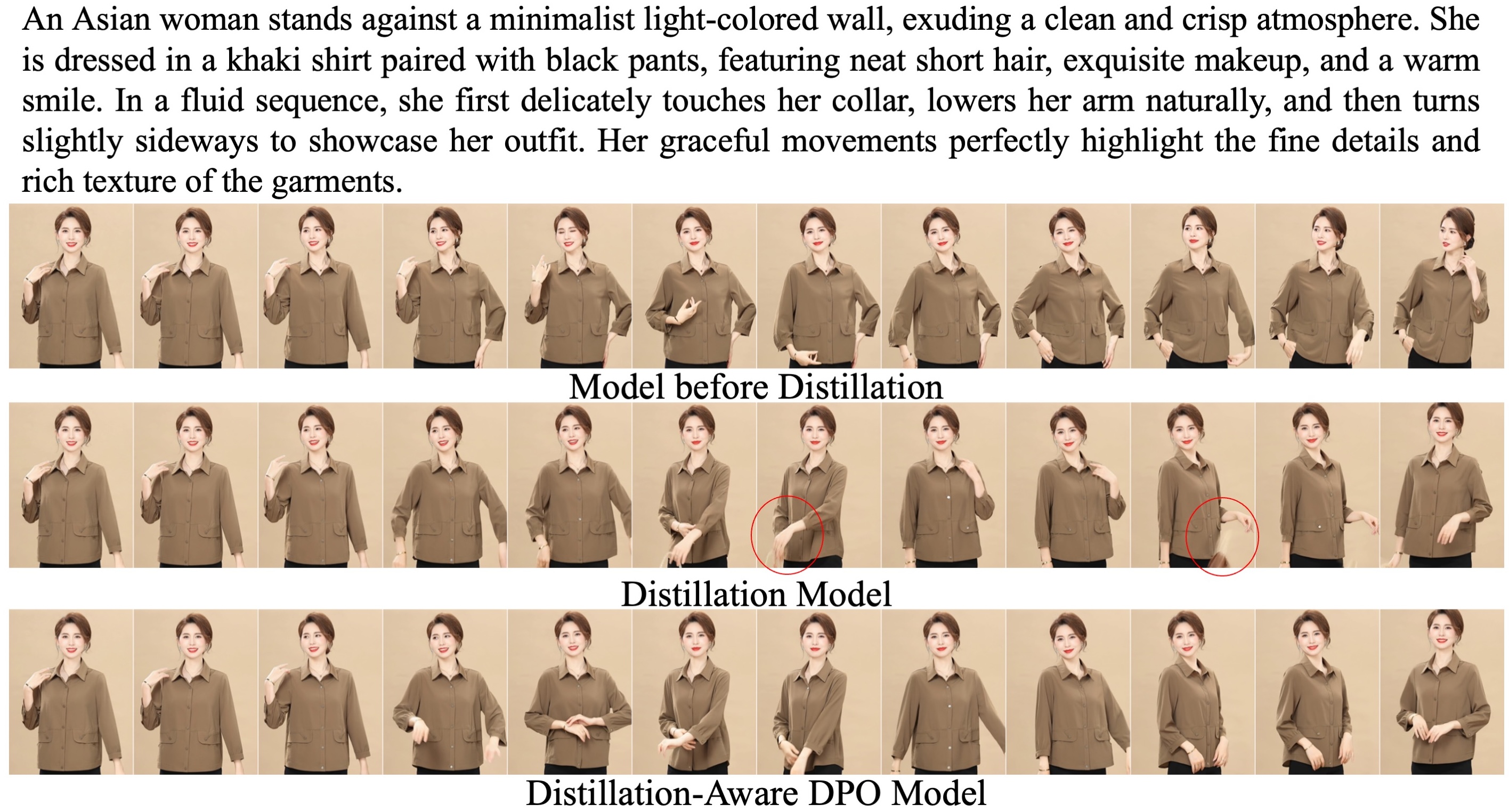}
    \caption{\textbf{Illustrative Example 2 of Post-Distillation Recovery and Enhancement through DPO.} The top row shows the original teacher model's high-fidelity output. The middle row presents the naively distilled student model, which exhibits visual degradation despite a $3 \times$ speedup. The bottom row displays our DPO-optimized model, successfully restoring and often surpassing the teacher's generative fidelity, thus proving DPO's ability to overcome distillation losses and enhance final output while maintaining efficiency.}
    \label{app_fig:vis32}
\end{figure}

\begin{figure}
    \centering
    \includegraphics[width=1.0\linewidth]{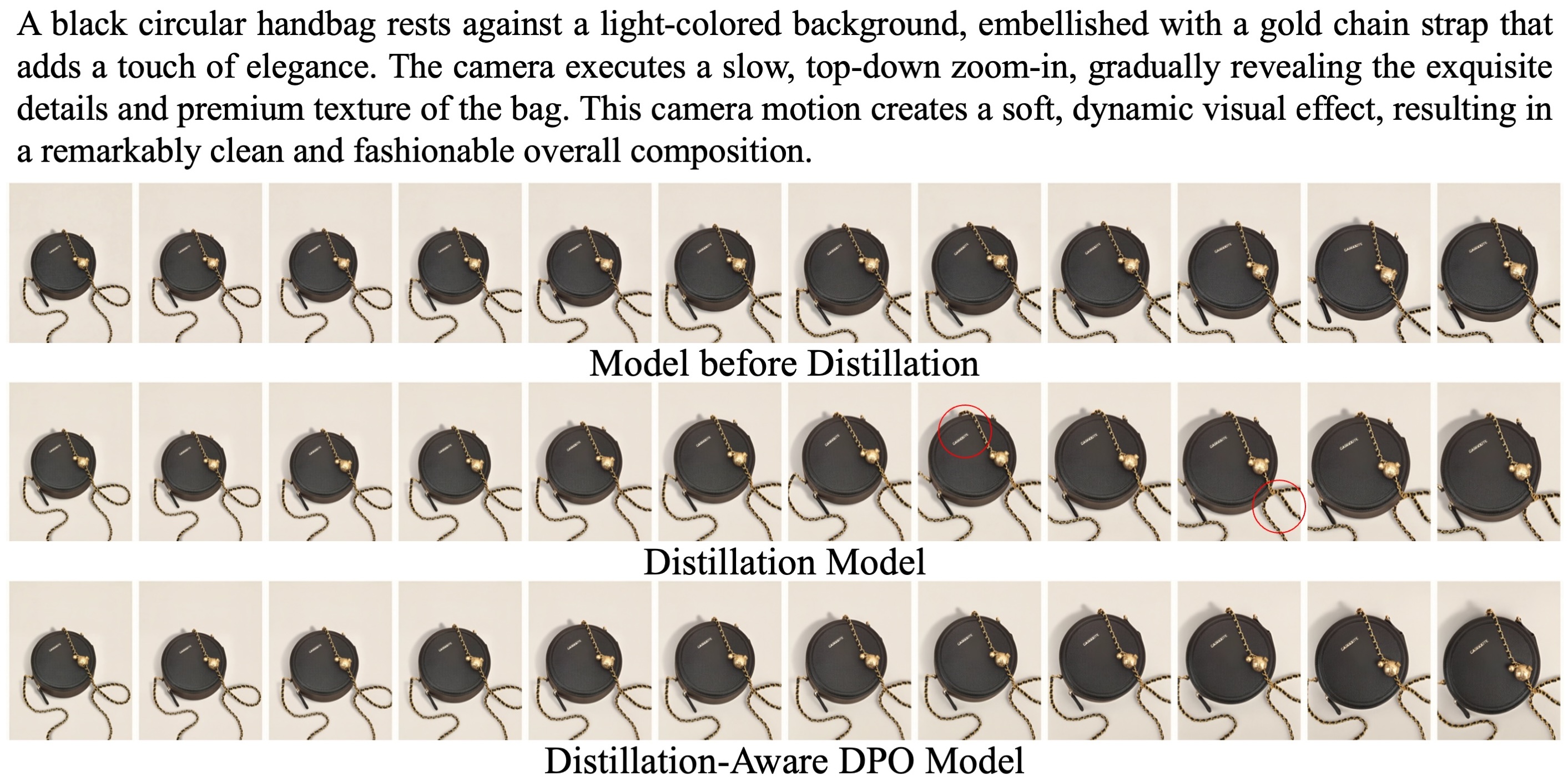}
    \caption{\textbf{Illustrative Example 3 of Post-Distillation Recovery and Enhancement through DPO.} The top row shows the original teacher model's high-fidelity output. The middle row presents the naively distilled student model, which exhibits visual degradation despite a $3 \times$ speedup. The bottom row displays our DPO-optimized model, successfully restoring and often surpassing the teacher's generative fidelity, thus proving DPO's ability to overcome distillation losses and enhance final output while maintaining efficiency.}
    \label{app_fig:vis33}
\end{figure}

(2) \textbf{Post-Distillation Recovery and Enhancement}: This section validates the necessity of our distillation-aware DPO phase. We compare the naively distilled model—which, despite providing a $3\times$ inference speedup, suffers from visual degradation—against our final optimized policy. As presented in Figs.~\ref{app_fig:vis31}, ~\ref{app_fig:vis32}, and~\ref{app_fig:vis33}, this comparison unequivocally illustrates the quality-speed trade-off inherent in standard distillation. Crucially, it confirms that our terminal optimization phase effectively reverses these degradation effects, restoring original generative fidelity and, in several instances, surpassing the pre-distillation baseline. This demonstrates that our framework attains an optimal balance between computational efficiency and superior visual quality.

%%%%%%%%%%%%%%%%%%%%%%%%%%%%%%%%%%%%%%%%%%%%%%%%%%%%%%%%%%%%

\end{document}